\documentclass[aps, prb, showpacs, amsmath, letterpaper]{revtex4-2}

\usepackage{graphics}
\usepackage{graphicx}

\usepackage{amssymb}
\usepackage{bm}

\usepackage[english]{babel}

\usepackage{tabularx}
\usepackage{booktabs}
 
\usepackage{amsmath}
\usepackage{graphicx}
\usepackage{wrapfig}

\usepackage{adjustbox}
\usepackage{mathtools}  

\usepackage[colorlinks=true, allcolors=blue]{hyperref}

\begin{document}

\title{Learning Beyond Optimization: Stress-Gated Dynamical Regime Regulation in Autonomous Systems}

\author{Sheng Ran}

\affiliation{Department of Physics, Washington University in St. Louis, St. Louis, MO 63130, USA}

\affiliation{Reconstructing Future Research Initiative}

\date{\today}

\begin{abstract}

Despite their apparent diversity, modern machine learning methods can be reduced to a remarkably simple core
principle: learning is achieved by continuously optimizing parameters to minimize or maximize a scalar objective
function. This paradigm has been extraordinarily successful for well-defined tasks where goals are fixed and evaluation criteria are explicit. However, if artificial systems are to move toward true autonomy—operating over long horizons and across evolving contexts—objectives may become ill-defined, shifting, or entirely absent. In such settings, a fundamental question emerges: in the absence of an explicit objective function, how can a system determine whether its ongoing internal dynamics are productive or pathological? And how should it regulate structural change without external supervision?
In this work, we propose a dynamical framework for learning without an explicit objective. Instead of minimizing external error signals, the system evaluates the intrinsic health of its own internal dynamics and regulates structural plasticity accordingly. We introduce a two-timescale architecture that separates fast state evolution from slow structural adaptation, coupled through an internally generated stress variable that accumulates evidence of persistent dynamical dysfunction. Structural modification is then triggered not continuously, but as a state-dependent event. Through a minimal toy model, we demonstrate that this stress-regulated mechanism produces temporally segmented, self-organized learning episodes without reliance on externally defined goals. Our results suggest a possible route toward autonomous learning systems capable of self-assessment and internally regulated structural reorganization.

\end{abstract}

\maketitle{}

Despite the remarkable achievements of modern artificial intelligence, a fundamental limitation remains: today’s AI systems are not truly autonomous. Across supervised~\cite{Kotsiantis,Singh2016,Nasteski2017,Jiang2020}, reinforcement~\cite{Jia2020,Shakya2023}, and self-supervised paradigms~\cite{Hastie,Rani2023,Shukla2025}, learning ultimately relies on the continuous optimization of an explicit loss or objective function defined—directly or indirectly—by human designers. While this paradigm has proven extraordinarily successful for well-specified tasks, it presupposes the existence of clearly defined goals and stable optimization landscapes. However, genuine open-ended learning often unfolds without predefined objectives. In biological evolution, human cognition, and creative discovery, goals frequently emerge only retrospectively. Agents do not begin with a fixed scalar objective; instead, they navigate uncertain, evolving, and partially defined environments where both tasks and representations may change over time.

Over the past decades, several approaches have attempted to move beyond externally specified task losses. Broadly, these efforts can be grouped into three categories. The first category replaces explicit task objectives with intrinsic scalar quantities~\cite{Linsker1988,Rao1999,Jaynes1957,Hinton2002}. Examples include mutual information maximization, prediction error minimization, entropy maximization, contrastive divergence, and variational free energy minimization. Although these methods do not depend on externally labeled targets, learning remains driven by the optimization of a scalar function defined over system states or representations. In this sense, the objective is reformulated rather than eliminated. The second category adopts local or biologically inspired update rules such as Hebbian learning~\cite{Oja1982}, self-organizing maps~\cite{Kohonen1982}, and spike-timing-dependent plasticity. These rules often appear objective-free at the algorithmic level. However, in many cases, their dynamics can be shown to correspond to the descent of an implicit energy or Lyapunov function under certain assumptions~\cite{Hopfield1982}. Even when no explicit loss is written, learning frequently remains governed by an underlying scalar potential structure. The third category, including the free energy principle and active inference frameworks, explicitly formalizes adaptation as the minimization of variational free energy~\cite{Friston2009}. Although philosophically ambitious, these approaches likewise retain optimization as the central organizing principle.

Taken together, current mainstream approaches to learning without external objectives typically replace task-specific losses with alternative scalar potentials. Learning remains largely characterized as optimization within a fixed state space under the influence of a scalar-driven dynamical field. For many engineering purposes—training models, solving well-defined tasks, or optimizing performance metrics—this paradigm is both effective and appropriate. However, if the goal is to construct truly open-ended, self-evolving cognitive systems, deeper questions arise.

Optimization-based learning implicitly assumes that:
\begin{itemize}
    \item a meaningful scalar objective exists,
    \item the representational space is sufficiently stable,
    \item and improvement corresponds to descending (or ascending) a fixed potential landscape.
\end{itemize}
Yet autonomous intelligence in biological systems exhibits additional properties: it maintains itself, reorganizes its internal structure, revises its representational assumptions, and escapes unproductive modes of reasoning—even in the absence of clearly defined objectives. Such systems do not merely optimize within a static landscape; they may alter the landscape itself.

This leads to a fundamental question: when no explicit loss function is available, how can a system determine whether its ongoing reasoning process is productive or pathological?
Current AI systems rarely confront this question because they are designed as tools rather than autonomous agents. Human operators define objectives, schedule training, reset models, switch datasets, and intervene when systems fail. Under this regime, AI does not need to evaluate the quality of its own reasoning dynamics—this responsibility is outsourced to humans. However, this assumption breaks down in settings characterized by long-horizon autonomy (e.g., autonomous scientific discovery or exploratory agents), continual task evolution, or open-ended creative exploration. In such environments, the central problem shifts from performance optimization to structural self-assessment: determining whether the system’s internal organization remains conducive to sustained reasoning.

In this work, we study how a learning system can evaluate and regulate its own reasoning dynamics without being continuously driven by a single scalar gradient objective. Rather than framing learning solely as parameter adjustment within a fixed landscape, we treat it as the regulation of dynamical regimes. We propose a learning framework that incorporates phase-gated, discrete, and structurally constrained plasticity. Instead of continuously optimizing a global scalar objective, the system monitors dynamical signatures of its own reasoning process and selectively triggers structural reorganization when pathological regimes are detected. We show that such a framework enables a system to regulate and reorganize potentially unproductive reasoning trajectories, even in the absence of an externally specified objective.

\begin{figure*}[ht]
    \centering
    \includegraphics[width=\linewidth]{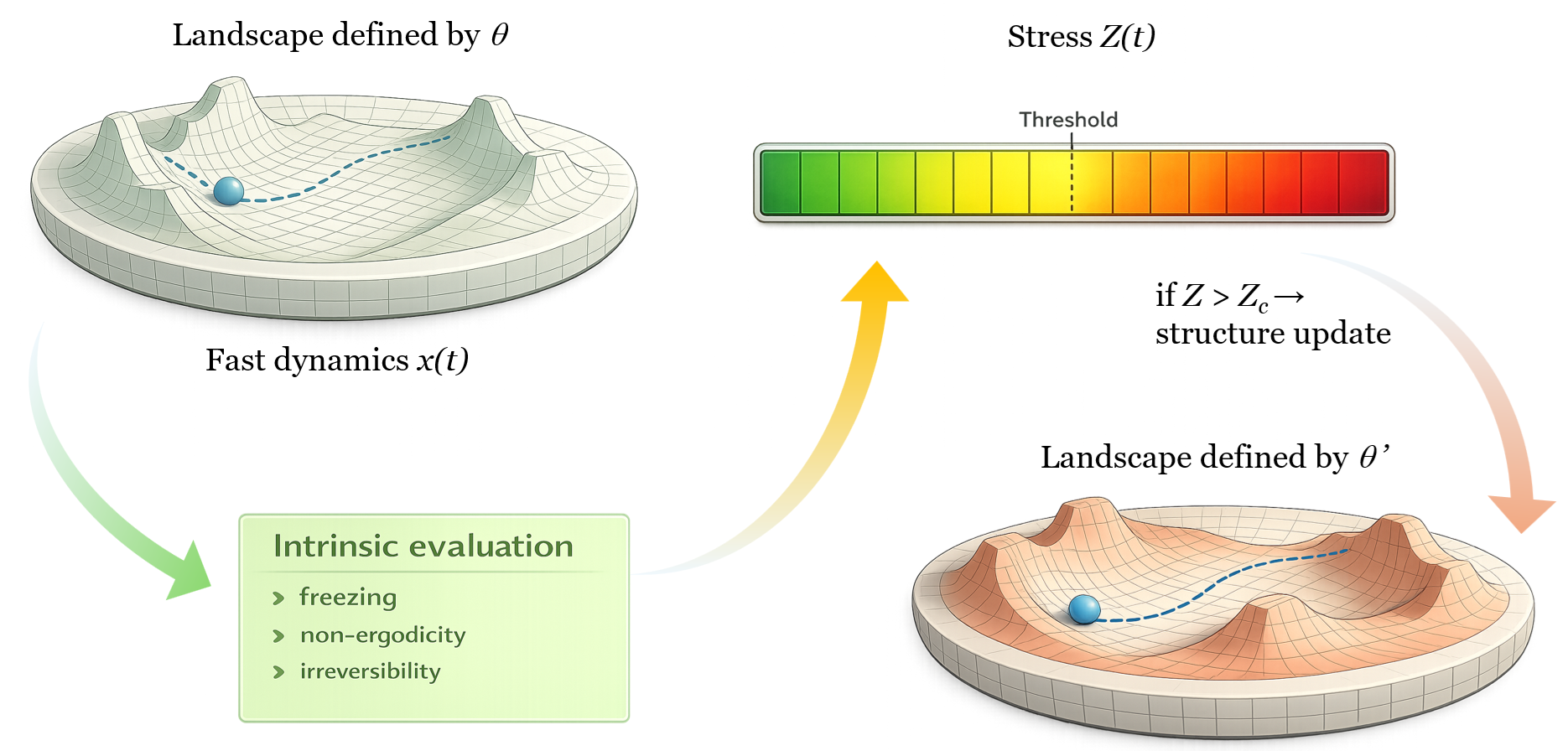}
    \caption{\textbf{Two-Timescale Dynamical Framework.} Fast dynamics $x(t)$ evolve within a structural landscape defined by slow parameters $\theta$. The system continuously performs intrinsic evaluation of its own dynamical health (e.g., freezing, non-ergodicity, irreversibility), generating an accumulated stress signal $Z(t)$.When stress exceeds a critical threshold $Z_c$, structural update is triggered, leading to a reconfiguration of the landscape $\theta \rightarrow \theta'$. The updated structure reshapes the effective landscape in which subsequent fast dynamics unfold.}
    \label{framework}
\end{figure*}

\section{A Two-Timescale Dynamical Framework}
The key question we address here is: instead of minimizing an externally defined error, can a system regulate its plasticity based on the intrinsic health of its own dynamics?
To investigate this possibility, we develop a minimal two-timescale dynamical framework in which fast state evolution and slow structural adaptation are explicitly separated.

In the presence of an explicit objective function, failure modes collapse into a single signal provided by the loss. In the absence of an explicit objective, however, this collapse no longer holds. A system may fail either because it has not explored sufficiently within a viable structure, or because the structure itself is incapable of supporting effective reasoning. In the latter case, thinking may continue indefinitely without producing meaningful progress. Crucially, these two failure modes are indistinguishable at the level of outcomes alone.

Distinguishing between them therefore requires evaluating the thinking process itself, rather than its results. Without external feedback or task-specific loss, the only available signal is the intrinsic behavior of the system’s own internal dynamics. For this reason, “thinking” must be modeled not as a static input–output mapping, but as a physical process unfolding in time.

To formalize this idea, we introduce a coupled dynamical framework with a fundamental separation of timescales as shown in Fig.~\ref{framework}. This separation reflects a core cognitive distinction: transient thought activity evolves rapidly within a given representational structure, while the structure itself changes slowly and only under specific conditions. Such a distinction allows the system to assess whether failures arise from the dynamics of thinking within a structure, or from the structure itself.

\subsection{System Equations and Multiscale Dynamics}

We define the cognitive architecture through two coupled dynamical components representing fast and slow variables of the system.

\begin{itemize}
    \item \textbf{Fast State (Thinking Variable).}  
    The variable $\mathbf{x}(t) \in \mathbb{R}^n$ represents the instantaneous state of ongoing thought, such as neural population activity or working memory configurations. Its dynamics describe how the system explores, evaluates, and transitions between mental states within a fixed representational structure. We model this process as an overdamped Langevin dynamics,
    \begin{equation}
        \dot{\mathbf{x}} = -\nabla_{\mathbf{x}} V(\mathbf{x}; \bm{\theta}) + \bm{\eta}(t),
    \end{equation}
    where $V(\mathbf{x}; \bm{\theta})$ defines an effective cognitive landscape shaped by the structural parameters $\bm{\theta}$, and $\bm{\eta}(t)$ represents intrinsic fluctuations or environmental noise.

    \item \textbf{Slow Structure (Structural Parameters).}  
    The variable $\bm{\theta}(t) \in \mathbb{R}^m$ encodes the persistent structural organization that shapes the landscape of thought, such as representational geometry or synaptic connectivity. Its evolution is modulated by a plasticity control signal $m(t)$, which regulates when and how structural modification occurs: 
    \begin{equation}
        \dot{\bm{\theta}} = m(t) \cdot \mathbf{g}(\mathbf{x}, \bm{\theta}).
    \end{equation}
    The specific form of $m(t)$—whether continuously active, state-dependent, or event-driven—will be treated as a design choice to be analyzed later.
\end{itemize}

This formulation establishes a bidirectional feedback loop. The structural parameters $\bm{\theta}$ determine the geometry of the cognitive landscape and thus constrain the trajectories of thought $\mathbf{x}(t)$ (downward causality). Conversely, the long-term statistical properties of these trajectories—rather than immediate outcomes—regulate whether the structure itself should be revised (upward causality). This separation enables the system to tolerate transient thinking failures while responding decisively to persistent structural inadequacy.

\subsection{The Cognitive Stress Field}

To prevent the system from overreacting to transient fluctuations or short-lived failures, structural modification should depend on temporally integrated information about the quality of thinking. To achieve this, we introduce an intrinsic Cognitive Stress Field, $Z(t)$, which mediates between fast dynamics and structural change.

The stress field $Z(t)$ is a latent dynamical variable that accumulates tension when the quality of the thinking process degrades over time. Unlike traditional loss functions in machine learning—which rely on task-specific labels or external objectives—the evolution of $Z(t)$ depends on intrinsic dynamical health, with an additional internal cost term that discourages excessive plasticity:
\[
\dot Z=\Phi(\mathcal Q(\cdot))+\underbrace{\Psi(m,\Delta \theta)}_{\text{plasticity cost}}-\gamma Z,
\]
Here, $\mathcal{Q}(\cdot)$ denotes a set of dynamical descriptors evaluated over a finite time window, such as freezing, non-ergodicity, or irreversibility. $\Psi(m,\Delta\bm{\theta})$ represents an intrinsic cost of plasticity, and $\gamma$ represents a dissipation rate that prevents unbounded stress accumulation.

Structural plasticity is then regulated through the interaction between the stress field and the control signal $m(t)$. In a simple threshold-based implementation, one may define,
\begin{equation}
    m(t) = \Theta\!\big(Z(t) - Z_c\big),
\end{equation}
where $\Theta(\cdot)$ is the Heaviside step function and $Z_c$ denotes a critical stress level.

However, this threshold rule should be understood as one possible realization rather than a mandatory choice. More generally, $m(t)$ may represent a continuous modulation function, a probabilistic trigger, or an event-based mechanism derived from the stress dynamics. The key principle is that structural adaptation is governed by internally accumulated evidence of persistent dynamical pathology, rather than by instantaneous errors or externally defined objectives.

With this construction, structural change becomes state-dependent: transient or recoverable thinking failures may dissipate without modifying the structure, while sustained degradation in dynamical health gradually accumulates stress and increases the likelihood of structural reorganization.

\section{Criteria for “Good Thinking": Dynamical Descriptors}

\begin{figure*}[ht]
    \centering
    \includegraphics[width=\linewidth]{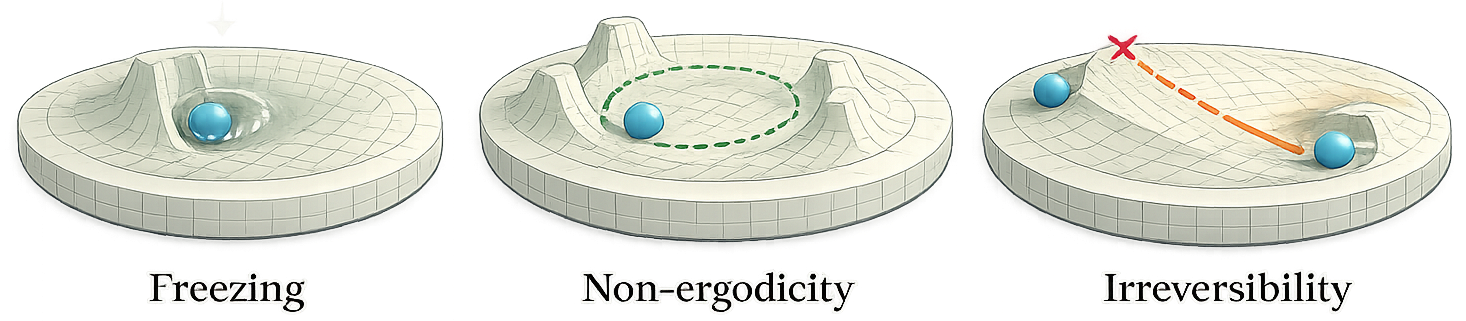}
    \caption{\textbf{Intrinsic dynamical failure modes.} (Left) Freezing. The system becomes confined to a narrow region of the landscape. (Middle) Non-ergodicity. The trajectory explores only a subset of the accessible landscape, failing to visit other basins over long timescales. (Right) Irreversibility. The trajectory undergoes a directed drift or structural bias such that previously accessible regions cannot be revisited.}
    \label{failure}
\end{figure*}

Having established a dynamical framework for thinking, we now face a central question: how can a system evaluate the quality of its own thinking in the absence of external objectives or task-specific feedback? In our framework, “good thinking” is not defined by performance, but by the intrinsic properties of the thinking dynamics themselves. We propose that “good thinking" is defined by the structural properties of the state-space trajectory $\mathbf{x}(t)$. To quantify the intrinsic health of these dynamics, we introduce three physically motivated metrics, which together govern the evolution of the cognitive stress field $Z(t)$, as shown in Fig.~\ref{failure}

\subsection{Freezing Index ($F_T$): Quantifying Attractor Collapse}

Persistent “looping" or “stagnation" in a narrow region of the state space is a hallmark of suboptimal cognition. To quantify this “early-freezing" phenomenon, we define the \textit{Freezing Index} based on the local covariance of the fast state:
\begin{equation}
    F_T = \exp\left( - \lambda \cdot \text{Tr}(\mathbf{\Sigma}_{\mathbf{x}[t-W:t]}) \right)
\end{equation}
where $\mathbf{\Sigma}$ is the covariance matrix of the trajectory calculated over a sliding window $W$, and $\lambda$ is a scaling sensitivity parameter. 
\begin{itemize}
    \item As the trajectory collapses into a point-like attractor or a low-dimensional limit cycle, $\text{Tr}(\mathbf{\Sigma}) \to 0$ and $F_T \to 1$, indicating “frozen thinking."
    \item A healthy thinking process maintains a sufficiently high trace, representing active exploration within the current landscape.
\end{itemize}

\subsection{Non-Ergodicity ($E_T$): Measuring Exploratory Scope}

While $F_T$ detects local collapse, a system may still be trapped within a suboptimal basin of attraction, failing to explore the broader relevant state space. We quantify this using the \textit{Non-Ergodicity} index, defined as the Kullback-Leibler (KL) divergence between the empirical occupancy distribution and a reference distribution:
\begin{equation}
    E_T = D_{KL}\left( p_T(\mathbf{x}) \parallel q(\mathbf{x}) \right)
\end{equation}
where $p_T(\mathbf{x})$ is the empirical distribution of states visited during the interval $T$, and $q(\mathbf{x})$ represents a target or uniform distribution over the reachable state space. A high $E_T$ value signifies that the thinking process is non-ergodic, remaining localized in a subset of the configuration space and ignoring potentially superior solutions.

\subsection{Irreversibility ($R_T$): Assessing Cognitive Flexibility}

Drawing from stochastic thermodynamics, we interpret the “flexibility" of thought through the lens of time-reversal symmetry. “Bad thinking" is often characterized by a “mental slide"—a strongly irreversible process where the system is driven into a state from which it cannot easily backtrack. We define the \textit{Irreversibility Index} as:
\begin{equation}
    R_T = \ln \frac{\mathcal{P}(\text{forward path})}{\mathcal{P}(\text{backward path})}
\end{equation}
where $\mathcal{P}$ denotes the path probability of the trajectory. In a computational context, this can be approximated by the discriminability between forward and time-reversed sequences of $\mathbf{x}$. 
\begin{itemize}
    \item $R_T \approx 0$ indicates near-reversible, “liquid" dynamics, suggesting a high capacity for error correction and logical(backtracking).
    \item High $R_T$ corresponds to high entropy production, signaling a brittle, “one-way" cognitive process that is prone to becoming trapped in “mental dead-ends".
\end{itemize}

\section{Continuous versus Stress-Gated Plasticity}

Having established an intrinsic, dynamical criterion for evaluating the quality of thinking—independent of any external task or loss—the next question concerns how a system should adapt when its internal dynamics degrade.
Within our framework, structural adaptation is regulated by the control signal $m(t)$. In principle, $m(t)$ may take different forms: it may be continuously active, weakly modulated, probabilistic, or discretely triggered. 

In the presence of an explicit loss function, continuous parameter optimization is natural: every update step is locally validated by the objective. Persistent adjustment is justified because the loss provides a reliable gradient signal that distinguishes improvement from deterioration at each moment. Indeed, in most contemporary learning paradigms, the implicit assumption is continuous plasticity. Formally, this corresponds to fixing
\[
m(t) \equiv 1,
\]
so that structural parameters evolve at every time step in response to moment-to-moment dynamics.

In the absence of an explicit objective, however, this justification disappears. Structural plasticity is no longer anchored to an externally defined direction of progress. The system must rely solely on internal dynamical signals, which are inherently noisy and fluctuating. Under such conditions, continuously active plasticity implicitly assumes that moment-to-moment deviations always warrant structural modification. This assumption is nontrivial.

Without an objective function to separate transient fluctuations from genuine structural inadequacy, always-on adaptation risks conflating two qualitatively different situations:
\begin{itemize}
    \item Local instability within an otherwise adequate structure.
    \item Persistent structural mismatch that cannot be resolved by internal exploration alone.
\end{itemize}

If plasticity is continuously active, these two cases are treated identically. Structural parameters are adjusted at every moment, even when short-term fluctuations may self-correct. In effect, structure is never allowed to remain stable long enough to be properly tested. The system cannot distinguish between “explore more within the current structure” and “change the structure itself.”

This observation suggests a more general principle: in the absence of an external objective, plasticity should not be permanently active. Instead, structural modification should require accumulated internal evidence that the current organization is persistently inadequate. Such evidence must integrate over time to avoid reacting to transient noise.

A gated or state-dependent plasticity mechanism naturally implements this requirement. By allowing structural updates only when an internally accumulated signal exceeds a threshold, the system creates a temporal separation between two phases:
\begin{itemize}
    \item Exploration within a fixed structure.
    \item Structural reorganization in response to sustained internal stress.
\end{itemize}

This separation is not introduced as a biological analogy, but as a logical consequence of operating without an explicit objective. When improvement cannot be externally validated at every step, structure must be granted periods of stability; otherwise, adaptation becomes indistinguishable from noise-driven drift.

\section{Stress-Gated Cognitive Dynamics}

We now construct a minimal toy model based on the principles developed above, which we term the Stress-Gated Cognitive Dynamics (SGCD) model. SGCD is designed to illustrate how high-dimensional internal states can spontaneously give rise to repeatable, low-dimensional behavioral patterns when the system is endowed with (i) a global stress-like variable and (ii) a gated plasticity mechanism that intermittently reshapes internal structure.

The model is not trained on external data and does not optimize an explicit objective function. Instead, it self-organizes by alternating between two modes: (a) fast intrinsic dynamics that explore the current structural landscape, and (b) discrete plasticity episodes that are triggered endogenously when accumulated stress signals persistent dynamical failure. Technically, SGCD is a discrete-time instantiation of the generic continuous-time framework introduced earlier, with the Langevin drift implemented via a contractive recurrent update and the stress field realized as an exponential moving average.

Before presenting the specific equations of the SGCD model, it is important to clarify how the general criteria for “good thinking” introduced in the theoretical section are instantiated in this minimal setting.

The freezing, non-ergodicity, and irreversibility indices defined in Sec.~II are intended as principle-level diagnostics. They formalize, at a conceptual level, what it would mean for an autonomous system to evaluate the intrinsic health of its own thinking dynamics without reference to external tasks or losses.

In the present toy model, we do not attempt to implement all three criteria in their full generality. Instead, we adopt a reduced and computationally tractable realization that captures the most salient failure modes relevant for spontaneous structural reorganization. Specifically, freezing is detected through relative stagnation in a noise-corrected velocity measure, while non-ergodicity is approximated by the absence of a stable prototype structure in the recent trajectory. The irreversibility criterion, while conceptually important, is not explicitly evaluated in this minimal model and is left for future extensions.

This reduction preserves the core logic of the framework while keeping the model analytically transparent. Stress is generated by persistent dynamical pathologies, rather than instantaneous fluctuations, and structural plasticity is gated only when fast exploration repeatedly fails to organize into coherent internal patterns.

\subsection{Microscopic state and fast dynamics}

The microscopic state is an $N$-dimensional vector $x(t)\in\mathbb{R}^N$, representing the system’s moment-to-moment cognitive state. It evolves under a recurrent interaction matrix $W(t)\in\mathbb{R}^{N\times N}$ with additive noise:
\[
x(t+1)=(1-\alpha)x(t)+\alpha\tanh\big(W(t)x(t)\big)+\sigma\eta(t),
\]
where $\alpha$ controls update speed and $\sigma$ sets the noise level.

The matrix $W(t)$ is symmetric with zero diagonal, and its scale is stabilized by enforcing a target spectral radius (details below). This prevents the fast dynamics from collapsing (vanishing interactions) or diverging (runaway amplification).

\subsection{Behavioral observables: motion and prototype strength}

At each time step, the model evaluates two windowed observables from the recent trajectory $X_{t-\tau:t}={x(t-\tau),\dots,x(t-1)}$:

\begin{itemize}
    \item Noise-corrected velocity. A velocity proxy measures how rapidly the state moves in configuration space:
\[
v_{\rm raw}(t)=\left\langle \frac{1}{N}|x(t+1)-x(t)|^2 \right\rangle,
\]
\[
v_{\rm eff}(t)=\max\big(v_{\rm raw}(t)-\sigma^2,0\big).
\]
A short-timescale EWMA produces a smoothed motion variable $v_{\rm smooth}(t)$.
    \item Prototype strength. A simple order parameter quantifies whether the trajectory has a consistent mean direction:
\[
Q(t)=\frac{|\mu(t)|}{\sqrt{N}},\quad \mu(t)=\langle x\rangle_{t-\tau:t}.
\]
Large $Q$ indicates the system is hovering around a biased prototype (more structured); small $Q$ indicates a more balanced or dispersed trajectory.
    \end{itemize}

\subsection{Badness and Stress accumulator $Z(t)$}

The model defines a scalar badness core $B_{\rm core}(t)\in[0,1]$ that reflects two conditions:

\begin{itemize}
    \item \textbf{Plateau / stagnation}: the system is moving unusually slowly compared to a lagged baseline of $v_{\rm eff}$.
    \item \textbf{Low prototype}: the system lacks strong prototype structure relative to a slow baseline of $Q$.
\end{itemize}

Concretely, a plateau score rises when $v_{\rm smooth}(t)$ falls below a fraction of a lagged baseline $v_{\rm base}(t)$, and a low-prototype factor rises when $Q(t)$ sits below an adaptive good level derived from a slow baseline $Q_{\rm base}(t)$.

The product forms the core badness:
\[
B_{\rm core}(t)\sim \text{Plateau}(t)\times \text{LowProto}(t),
\]
with optional flooring.

This construction makes badness a functional diagnostic: the system is not merely noisy or slow; it is slow and not forming a stable prototype.

A global stress variable $Z(t)$ integrates total badness (including plasticity costs) on a slow timescale:
\[
Z(t+1)=(1-\lambda_Z)Z(t)+\lambda_ZB_{\rm total}(t).
\]
Stress therefore encodes recent history and can remain elevated even if instantaneous badness fluctuates.

\subsection{Gated plasticity}

Plasticity is not always on. Instead, the model opens a plasticity episode when stress exceeds a threshold:

\begin{itemize}
    \item Normal gating: when the system is armed and $Z>Z_{\rm on}$.
    \item Override trigger: if stress remains high and fails to improve over a long window, the system can open a gate even without being armed, preventing prolonged stagnation under elevated stress.
    \item The system becomes re-armed only after stress falls below $Z_{\rm off}$ (hysteresis).
\end{itemize}

A gate triggers a commit window of length $L_{\rm commit}$, during which plasticity is ON at every step. After a gate, a refractory window prevents immediate retriggering.

This creates discrete cognitive events: short plastic episodes separated by longer periods of stable fast dynamics.

\begin{figure*}[ht]
    \centering
    \includegraphics[width=\linewidth]{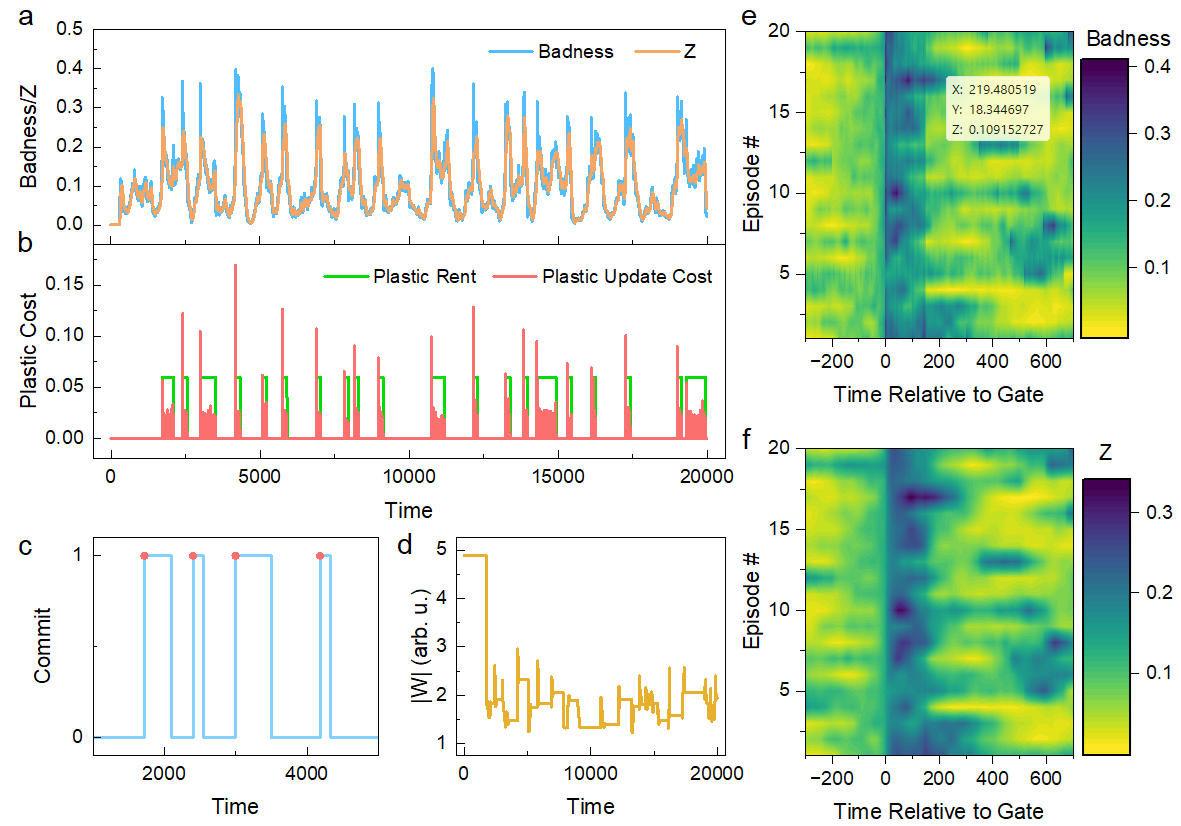}
    \caption{\textbf{Stress-gated cognitive dynamics.} (a) Badness and stress accumulator $Z$ as a function of the global time. (b) Plastic rent and plastic update cost as a function of the global time. (c) A zoomed view of the gating signal. The red dots represent the gate trigger. (d) The norm of the connectivity matrix $|W|$ as a function of the global time. (e) Gate-aligned heatmaps of badness. The horizontal axis denotes time relative to the gate, and the vertical axis indexes individual episodes. (f) Gate-aligned heatmaps of stress.}
    \label{SGCD}
\end{figure*}

\subsection{Structural proposal and plasticity update}

When a gate opens, the model proposes a structural target $W_{\rm target}$ from the recent trajectory window. It uses the demeaned covariance:
\[
C=\frac{1}{\tau}X_c^\top X_c,\quad X_c=X-\langle X\rangle,
\]
then symmetrizes, removes the diagonal, optionally soft-thresholds small entries, and critically rescales the matrix to a fixed spectral radius $\rho_{\rm target}$. This normalization is the key stability fix: it prevents the learning target from collapsing toward zero magnitude.

During plasticity, the system performs a small convex step toward the target:
\[
W \leftarrow (1-\epsilon)W+\epsilon W_{\rm target},
\]
followed by symmetry/zero-diagonal enforcement and spectral-radius normalization.

During a plasticity episode, $W_{\rm target}$ may be refreshed every $L_{\rm refresh}$ steps using the most recent trajectory window $X_{t-\tau:t}$, allowing the structural update to reflect changes in the fast dynamics over the course of the episode.

\subsection{Plasticity has a price}

Plasticity is not free. The model includes two explicit costs:

\begin{itemize}
    \item Rent cost $c_{\rm on}$: a constant penalty paid whenever plasticity is ON.
    \item Update cost $c_W$: proportional to the relative RMS change of $W$ per step:
\[
   \text{cost}_{\rm update}\propto c_W\frac{{\rm RMS}(W_{\rm new}-W_{\rm old})}{{\rm RMS}(W_{\rm old})}.
\]
\end{itemize}

Total badness is:
\[
B_{\rm total}(t)=\text{clip}\Big(B_{\rm core}(t)+\text{rent}(t)+\text{updateCost}(t),0,1\Big).
\]
Because stress integrates $B_{\rm total}$, this creates a tradeoff: plasticity can reduce future badness, but it temporarily increases badness via costs—so the system cannot benefit from plasticity unless structural changes compensate for this price.

\subsection{Two safety mechanisms: early-abort and forced rearm}

To avoid two failure modes (wasting plasticity, or becoming permanently static after one ineffective gate), the model adds:

\begin{itemize}
    \item Early-abort (within a gate). Inside a plasticity episode, after a short delay $L_{\rm abort}$, the model measures the stress drop since gate onset, $\Delta Z= Z(t_{gate})-Z(t)$. If $\Delta Z$ is below a threshold while stress remains above $Z_{on}$ (and, optionally, if badness has not improved), the plasticity window is terminated early. This prevents long, unproductive plasticity runs.
    \item Forced rearm probes. If a gate is aborted for insufficient improvement, the model schedules a forced rearm: after cooldown it tries a shorter “probe” gate of length $L_{\rm probe}$, up to a maximum number of attempts per episode. This prevents the system from entering a state where it never gates again simply because the earlier gate did not succeed. 
\end{itemize}

\begin{figure*}[ht]
    \centering
    \includegraphics[width=\linewidth]{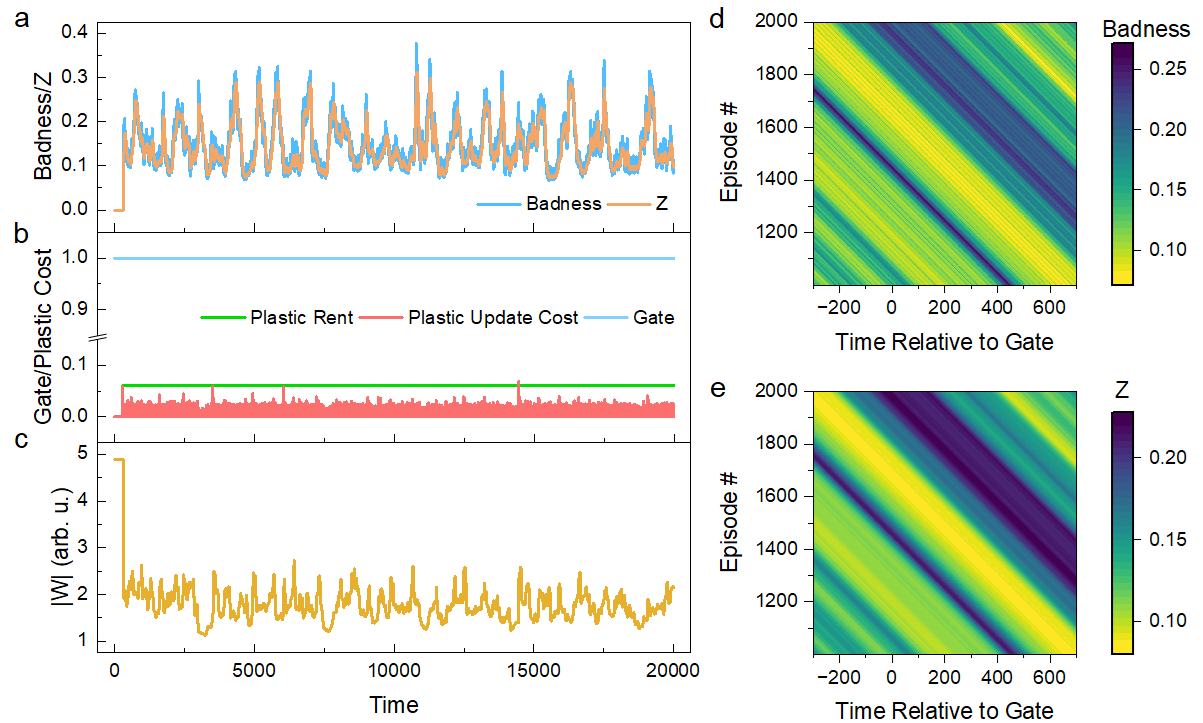}
    \caption{\textbf{Continuous Plasticity Dynamics.} (a) Badness and stress accumulator $Z$ as a function of the global time. (b) Gate, plastic rent and plastic update cost as a function of the global time. (c) The norm of the connectivity matrix $|W|$ as a function of the global time. (d) Gate-aligned heatmaps of badness. The horizontal axis denotes time relative to the gate, and the vertical axis indexes individual episodes. (e) Gate-aligned heatmaps of stress.}
    \label{continous}
\end{figure*}

\subsection{Results and Discussion}

Our results are summarized in Fig.~1. The global time series of total badness $B(t)$ and stress $Z(t)$ exhibit repeated cycles of stress accumulation followed by relaxation (Fig.~\ref{SGCD}a and Fig.~\ref{SGCD}b). Crucially, these relaxation phases are temporally aligned with discrete gate events that activate structural plasticity. Between gates, the system evolves solely under fast intrinsic dynamics and noise, during which stress gradually builds up. This behavior demonstrates that adaptation in the SGCD model is not continuous, but event-driven: structural updates are sparse and are triggered only when accumulated stress exceeds a critical threshold.

A closer view of the gating signal (Fig.~\ref{SGCD}c) shows that each gate initiates a finite commit window during which plasticity is active. The duration of these windows is bounded and independent of the instantaneous stress level once the gate is triggered. This design enforces a clear separation of time scales: fast dynamics explore the state space under a fixed structure, while slow structural plasticity operates episodically. Early-abort and refractory mechanisms further prevent runaway or redundant updates, ensuring that plasticity remains selective and economical.

To determine whether gate events merely coincide with random fluctuations or correspond to reproducible dynamical transitions, we align trajectories around gate onset and visualize them as heatmaps (Fig.~\ref{SGCD}e and Fig.~\ref{SGCD}f). If gates were triggered by incidental fluctuations, alignment would not reveal any systematic structure: individual episodes would remain heterogeneous and no consistent pre–post pattern would emerge. Instead, we observe a stereotyped temporal profile across events. Both badness and stress reliably peaks near gate onset and subsequently decays over hundreds of steps. Although individual episodes are not identical, they share a common dynamical motif. This alignment therefore reveals something nontrivial: gate onset defines an internally generated event time that organizes the dynamics into repeatable episodes. The system does not simply fluctuate continuously under noise; it alternates between accumulation phases and relaxation phases with a consistent temporal structure. In this sense, gates do not merely label high-stress moments—they segment the ongoing dynamics into coherent transitions between metastable regimes.

Finally, the norm of the connectivity matrix $|W|$ remains bounded over long times (Fig.~\ref{SGCD}d), but more importantly, it exhibits a piecewise-stable pattern: extended plateaus of nearly constant magnitude interrupted by discrete structural shifts. During plateau phases, plasticity is inactive and the structural landscape remains effectively frozen, allowing the fast dynamics to unfold within a stable representational substrate. Structural changes occur only during brief gated episodes, after which the system settles into a new metastable configuration. This punctuated pattern of structural evolution is nontrivial. Although plasticity is repeatedly invoked, the system does not drift continuously. Instead, it alternates between consolidation phases and reorganization phases, producing a temporally segmented structural trajectory rather than a diffusive one. The resulting connectivity dynamics therefore reflect episodic restructuring rather than gradual erosion or unstructured fluctuation.

To isolate the role of stress-gated plasticity, we perform a control experiment in which plasticity is continuously active at all times, while all other components of the model are kept identical.

Under continuous plasticity, the system remains dynamically stable: stress and badness do not diverge, and the connectivity norm stays bounded over long times (Fig.~\ref{continous}). Transient fluctuations and recurring peaks are clearly present in the global time series, indicating that continuous plasticity is sufficient to support ongoing adaptation and short-lived pattern formation. In this sense, the system remains responsive and dynamically active.

However, the qualitative organization of these dynamics differs fundamentally from the gated case. Although stress and badness exhibit repeated peaks that may visually resemble gated episodes, these fluctuations arise from ongoing noise-driven perturbations coupled to continuous weight updates. They do not correspond to discrete structural transitions. The system remains in a state of persistent drift rather than alternating between consolidation and reorganization phases.

This difference becomes clearer when examining the structural trajectory. Unlike the gated system, the connectivity norm $|W|$ does not exhibit extended plateaus separated by abrupt shifts. Instead, it fluctuates continuously without identifiable metastable regimes (Fig.~\ref{continous}c). The absence of structural plateaus indicates that no stable internal configuration is ever temporarily consolidated; plasticity is constantly modifying the substrate on which fast dynamics unfold.

To further probe whether reproducible learning episodes emerge, we apply the same gate-centered alignment analysis used in the gated system (Fig.~\ref{continous}d and Fig.~\ref{continous}e). Because plasticity is always active, each time step can be treated as a putative “episode onset,” and we align trajectories accordingly. In contrast to the gated case, no reproducible event-centered temporal structure emerges. The aligned heatmaps exhibit diagonal stripe patterns, reflecting the underlying global oscillatory fluctuations of the system. However, these patterns are not anchored to the alignment point. Instead, they represent phase-drifting dynamics that persist independently of the chosen reference time. When averaged across aligned segments, no consistent pre–post transition centered at the putative onset appears. This result indicates that, although continuous plasticity sustains ongoing fluctuations and transient structure formation, it does not organize the dynamics into phase-locked learning events. The system remains temporally stationary and homogeneous rather than episodically segmented. Learning unfolds as a continuous drift rather than as internally demarcated transitions.

Taken together, these results demonstrate that the SGCD model does more than evolve a collection of vectors under noise. The system self-organizes into a regime of punctuated adaptation, in which internally generated stress gates discrete episodes of structural plasticity. These gates segment the dynamics into reproducible learning events, rather than allowing adaptation to proceed diffusely in time. Importantly, this organization emerges without any externally imposed task, reward signal, or objective function. The system does not converge to a predefined attractor or optimize a fixed loss; instead, it develops a temporal structure of exploration, stress accumulation, and structural reconfiguration. In this sense, what is “learned” is not a static weight configuration, but a repeatable dynamical motif that governs when and how structural change occurs. SGCD therefore provides a minimal yet concrete example of how meaningful temporal organization—and episodic learning—can arise from unguided dynamics through stress-regulated plasticity alone.

\section{Conclusion and Outlook}

In this work, we have proposed a minimal dynamical framework for learning in the absence of an explicit objective function. Rather than treating learning as continuous optimization of a predefined loss, we model it as a two-timescale process in which fast state evolution unfolds within a structural landscape, while slow structural change is regulated by intrinsic evaluation of the system’s own dynamical health. Plasticity is not assumed to be perpetually active, but becomes a state-dependent phase triggered by accumulated internal stress.

This shift reframes the problem of learning. Instead of asking how to minimize error, we ask how a system can remain dynamically viable when no external criterion is available. The central claim is that meaningful structural organization can emerge not from task-specific supervision, but from the regulation of internal failure modes such as freezing, non-ergodicity, and irreversibility.

Although the present model is intentionally minimal, it suggests several broader implications. First, the framework provides a principled route toward genuinely autonomous systems. As artificial agents move beyond closed tasks and externally curated training regimes, the assumption of a fixed objective function becomes increasingly untenable. Long-horizon exploration, scientific discovery, creative reasoning, and continual adaptation all involve phases in which goals are unknown or only retrospectively defined. In such regimes, the capacity to evaluate and regulate one’s own dynamical organization may become more fundamental than objective optimization itself.

Second, the separation between fast and slow dynamics offers a structural perspective on stability and plasticity. Rather than tuning a continuous learning rate, the system organizes learning into phases of exploration and phases of reconfiguration. This temporal segmentation may provide a foundation for developing architectures that preserve internal coherence over long durations while remaining capable of decisive structural change.

Third, the stress-regulated mechanism opens the door to new mathematical questions. What classes of intrinsic dynamical metrics are sufficient to detect structural inadequacy? Under what conditions does gated plasticity produce stable nontrivial structure rather than drift or collapse? How do different stress accumulation rules affect long-term organization? These questions point toward a broader theory of self-regulated dynamical systems.

Several extensions naturally follow. The current framework can be generalized to high-dimensional neural networks, to stochastic control settings, or to agents embedded in partially observable environments. It may also be connected to biological theories of neuromodulation, sleep-dependent consolidation, and developmental critical periods, where structural change is episodic rather than continuous.

More fundamentally, this work suggests a conceptual distinction between optimization-driven learning and viability-driven learning. The former assumes a predefined target; the latter assumes only the necessity of maintaining coherent internal dynamics. Whether such viability-based regulation can support open-ended intelligence remains an open question. The present framework provides a minimal testbed for exploring that possibility.

\bibliographystyle{unsrt}
\bibliography{AI}

@article{Kotsiantis, title={Supervised Machine Learning: A Review of Classification Techniques}, volume={31}, url={https://www.informatica.si/index.php/informatica/article/view/148}, abstractNote={Supervised Machine Learning: A Review of Classification Techniques}, number={3}, journal={Informatica}, author={Kotsiantis, S.B.} }

@Article{Jiang2020,
  author                 = {Jiang, Tammy and Gradus, Jaimie L. and Rosellini, Anthony J.},
  journal                = {Behavior therapy},
  title                  = {Supervised Machine Learning: A Brief Primer.},
  year                   = {2020},
  month                  = {Sep},
  pages                  = {675-687},
  volume                 = {51},
  address                = {England},
  article-doi            = {10.1016/j.beth.2020.05.002},
  article-pii            = {S0005-7894(20)30067-8},
  completed              = {20210204},
  electronic-issn        = {1878-1888},
  electronic-publication = {20200516},
  grantno                = {R21 MH119492/MH/NIMH NIH HHS/United States},
  history                = {2021/09/01 00:00 [pmc-release]},
  issue                  = {5},
  language               = {eng},
  linking-issn           = {0005-7894},
  location-id            = {10.1016/j.beth.2020.05.002 [doi]},
  manuscript-id          = {NIHMS1595274},
  nlm-unique-id          = {1251640},
  print-issn             = {0005-7894},
  publication-status     = {ppublish},
  revised                = {20240329},
  source                 = {Behav Ther. 2020 Sep;51(5):675-687. doi: 10.1016/j.beth.2020.05.002. Epub 2020 May 16.},
  status                 = {MEDLINE},
  subset                 = {IM},
  termowner              = {NOTNLM},
  title-abbreviation     = {Behav Ther},
}

@InProceedings{Singh2016,
  author    = {Singh, Amanpreet and Thakur, Narina and Sharma, Aakanksha},
  booktitle = {2016 3rd International Conference on Computing for Sustainable Global Development (INDIACom)},
  title     = {A review of supervised machine learning algorithms},
  year      = {2016},
  pages     = {1310-1315},
}

@Article{Nasteski2017,
  author  = {Vladimir Nasteski},
  journal = {Horizons},
  title   = {An overview of the supervised machine learning methods},
  year    = {2017},
  pages   = {51-62},
  volume  = {4},
  url     = {https://api.semanticscholar.org/CorpusID:171520859},
}

@Book{Hastie,
  author    = {Trevor Hastie, Jerome Friedman, Robert Tibshirani},
  publisher = {Springer New York, NY},
  title     = {The Elements of Statistical Learning, Data Mining, Inference, and Prediction},
  year      = {2013},
  edition   = {1},
  isbn      = {978-0-387-21606-5},
  series    = {Springer Series in Statistics},
  doi       = {https://doi.org/10.1007/978-0-387-21606-5},
}

@Article{Shakya2023,
  author   = {Shakya, Ashish Kumar and Pillai, Gopinatha and Chakrabarty, Sohom},
  journal  = {Expert Systems with Applications},
  title    = {Reinforcement learning algorithms: A brief survey},
  year     = {2023},
  issn     = {0957-4174},
  pages    = {120495},
  volume   = {231},
  doi      = {10.1016/j.eswa.2023.120495},
  url      = {https://www.sciencedirect.com/science/article/pii/S0957417423009971},
}

@InProceedings{Jia2020,
  author    = {Jia, Jingkai and Wang, Wenlin},
  booktitle = {2020 35th Youth Academic Annual Conference of Chinese Association of Automation (YAC)},
  title     = {Review of reinforcement learning research},
  year      = {2020},
  pages     = {186-191},
  doi       = {10.1109/YAC51587.2020.9337653},
}

@Article{Rani2023,
  author                 = {Rani, Veenu and Nabi, Syed Tufael and Kumar, Munish and Mittal, Ajay and Kumar, Krishan},
  journal                = {Archives of computational methods in engineering : state of the art reviews},
  title                  = {Self-supervised Learning: A Succinct Review.},
  year                   = {2023},
  pages                  = {2761-2775},
  volume                 = {30},
  address                = {Netherlands},
  article-doi            = {10.1007/s11831-023-09884-2},
  article-pii            = {9884},
  electronic-issn        = {1886-1784},
  electronic-publication = {20230120},
  history                = {2023/01/20 00:00 [pmc-release]},
  issue                  = {4},
  language               = {eng},
  linking-issn           = {1134-3060},
  location-id            = {10.1007/s11831-023-09884-2 [doi]},
  nlm-unique-id          = {101728768},
  print-issn             = {1134-3060},
  publication-status     = {ppublish},
  revised                = {20230417},
  source                 = {Arch Comput Methods Eng. 2023;30(4):2761-2775. doi: 10.1007/s11831-023-09884-2. Epub 2023 Jan 20.},
  status                 = {PubMed-not-MEDLINE},
  termowner              = {NOTNLM},
  title-abbreviation     = {Arch Comput Methods Eng},
}

@InProceedings{Shukla2025,
  author    = {Shukla, Shristi and Dachawar, Madhavi and Logade, Harshal and Kadu, Sana},
  booktitle = {2025 International Conference on Computational, Communication and Information Technology (ICCCIT)},
  title     = {Self-Supervised Learning: The Core of Next-Gen Machine Learning and a Paradigm Shift in AI},
  year      = {2025},
  pages     = {486-491},
  doi       = {10.1109/ICCCIT62592.2025.10928142},
}

@Article{Linsker1988,
  author   = {Linsker, R.},
  journal  = {Computer},
  title    = {Self-organization in a perceptual network},
  year     = {1988},
  number   = {3},
  pages    = {105-117},
  volume   = {21},
  doi      = {10.1109/2.36},
}

@Article{Rao1999,
  author   = {Rao, Rajesh P. N. and Ballard, Dana H.},
  journal  = {Nature Neuroscience},
  title    = {Predictive coding in the visual cortex: a functional interpretation of some extra-classical receptive-field effects},
  year     = {1999},
  issn     = {1546-1726},
  number   = {1},
  pages    = {79--87},
  volume   = {2},
  doi      = {10.1038/4580},
  refid    = {Rao1999},
  url      = {https://doi.org/10.1038/4580},
}

@Article{Jaynes1957,
  author    = {Jaynes, E. T.},
  journal   = {Phys. Rev.},
  title     = {Information Theory and Statistical Mechanics},
  year      = {1957},
  month     = {May},
  pages     = {620--630},
  volume    = {106},
  doi       = {10.1103/PhysRev.106.620},
  issue     = {4},
  numpages  = {0},
  publisher = {American Physical Society},
  url       = {https://link.aps.org/doi/10.1103/PhysRev.106.620},
}

@Article{Hinton2002,
  author  = {Hinton, Geoffrey E.},
  journal = {Neural Computation},
  title   = {Training Products of Experts by Minimizing Contrastive Divergence},
  year    = {2002},
  number  = {8},
  pages   = {1771-1800},
  volume  = {14},
  doi     = {10.1162/089976602760128018},
}

@Article{Oja1982,
  author   = {Oja, Erkki},
  journal  = {Journal of Mathematical Biology},
  title    = {Simplified neuron model as a principal component analyzer},
  year     = {1982},
  issn     = {1432-1416},
  number   = {3},
  pages    = {267--273},
  volume   = {15},
  doi      = {10.1007/BF00275687},
  refid    = {Oja1982},
  url      = {https://doi.org/10.1007/BF00275687},
}

@Article{Kohonen1982,
  author   = {Kohonen, Teuvo},
  journal  = {Biological Cybernetics},
  title    = {Self-organized formation of topologically correct feature maps},
  year     = {1982},
  issn     = {1432-0770},
  number   = {1},
  pages    = {59--69},
  volume   = {43},
  doi      = {10.1007/BF00337288},
  refid    = {Kohonen1982},
  url      = {https://doi.org/10.1007/BF00337288},
}

@Article{Hopfield1982,
  author             = {Hopfield, J. J.},
  journal            = {Proceedings of the National Academy of Sciences of the United States of America},
  title              = {Neural networks and physical systems with emergent collective computational abilities.},
  year               = {1982},
  month              = {Apr},
  pages              = {2554-8},
  volume             = {79},
  electronic-issn    = {1091-6490},
  linking-issn       = {0027-8424},
  print-issn         = {0027-8424},
  address            = {United States},
  article-doi        = {10.1073/pnas.79.8.2554},
  comment            = {Trends Neurosci. 2020 Jul;43(7):453-455. doi: 10.1016/j.tins.2020.04.002. PMID: 32386741},
  completed          = {19820826},
  history            = {1982/10/01 00:00 [pmc-release]},
  issue              = {8},
  language           = {eng},
  nlm-unique-id      = {7505876},
  publication-status = {ppublish},
  revised            = {20220408},
  source             = {Proc Natl Acad Sci U S A. 1982 Apr;79(8):2554-8. doi: 10.1073/pnas.79.8.2554.},
  status             = {MEDLINE},
  subset             = {IM},
  title-abbreviation = {Proc Natl Acad Sci U S A},
}

@Article{Friston2009,
  author   = {Friston, Karl},
  journal  = {Trends in Cognitive Sciences},
  title    = {The free-energy principle: a rough guide to the brain?},
  year     = {2009},
  issn     = {1364-6613},
  number   = {7},
  pages    = {293--301},
  volume   = {13},
  doi      = {10.1016/j.tics.2009.04.005},
  url      = {https://www.sciencedirect.com/science/article/pii/S136466130900117X},
}

\clearpage

\end{document}